\documentclass{ecai} 
\newcommand{\forreviewers}[1]{#1}

\usepackage{latexsym}
\usepackage{amssymb}
\usepackage{amsmath}
\usepackage{amsthm}
\usepackage{booktabs}
\usepackage{enumitem}
\usepackage{graphicx}
\usepackage{color}
\usepackage[dvipsnames]{xcolor}
\usepackage{textgreek}
\usepackage{float}
\usepackage{multirow}

\usepackage[]{todonotes}

\newcommand{\com}[1]{}

\usepackage[linesnumbered,ruled,vlined]{algorithm2e}
\DontPrintSemicolon

\SetCommentSty{mycommfont}

\usepackage[appendix=append,bibliography=common]{apxproof}

\newtheoremrep{theorem}{Theorem}
\newtheorem{lemma}[theorem]{Lemma}

\newtheorem{proposition}[theorem]{Proposition}

\newtheorem{definition}{Definition}
\usepackage{thmtools}
\declaretheorem{example}
\renewcommand{\thmcontinues}[1]{continued}

\newcommand{\BibTeX}{B\kern-.05em{\sc i\kern-.025em b}\kern-.08em\TeX}

\newcommand{\agents}{\mathcal{A}}
\newcommand{\candidates}{\mathcal{C}}
\newcommand{\n}{n}
\newcommand{\m}{m}
\newcommand{\Profile}{\mathcal{P}}
\newcommand{\PW}[2]{\text{PW}_{#1}(#2)}
\newcommand{\NW}[2]{\text{NW}_{#1}(#2)}
\newcommand{\smo}[3]{\delta_{#1}^{#2}(#3)}
\newcommand{\movo}[2]{\Delta_{#1}^{#2}}
\newcommand{\sm}[2]{\delta_{#1}(#2)}
\newcommand{\mov}[1]{\Delta_{#1}}

\newcommand{\Se}{\mathcal{S}}
\newcommand{\X}{\mathcal{X}}
\newcommand{\Y}{\mathcal{Y}}
\newcommand{\Yb}{\bar{\mathcal{Y}}}
\newcommand{\R}{\mathcal{R}}

\newcommand{\newAXp}{iAXp}
\newcommand{\iAXp}{\text{\textit{\newAXp }}}
\newcommand{\iAXps}{\text{\textit{\newAXp s }}}

\newcommand{\shortAXp}{SiAXp}
\newcommand{\shortAXps}{\shortAXp s}

\newcommand{\CXp}{\text{\textit{CXp }}}
\newcommand{\CXps}{\text{\textit{CXps }}}
\newcommand{\Map}{\text{\textit{Map}}}

\newcommand{\nulle}{null}
\DeclareMathOperator{\Borda}{Borda}
\DeclareMathOperator{\argmin}{argmin}

\begin{document}


\begin{frontmatter}


\paperid{2248} 


\title{Abductive and Contrastive Explanations\\ for Scoring Rules in Voting}


\author{\fnms{Clément}~\snm{Contet}}
\author{\fnms{Umberto}~\snm{Grandi}}
\author{\fnms{Jérôme}~\snm{Mengin}}


\address{IRIT, Université de Toulouse\\ \{clement.contet, umberto.grandi, jerome.mengin\}@irit.fr}


\begin{abstract}
We view voting rules as classifiers that assign a winner (a class) to a profile of voters' preferences (an instance). 
We propose to apply techniques from formal explainability, most notably abductive and contrastive explanations, to identify minimal subsets of a preference profile that either imply the current winner or explain why a different candidate was not elected. 
Formal explanations turn out to have strong connections with classical problems studied in computational social choice such as bribery, possible and necessary winner identification, and preference learning.
We design algorithms for computing abductive and contrastive explanations for scoring rules. For the Borda rule, we find a lower bound on the size of the smallest abductive explanations, and we conduct simulations to identify correlations between properties of preference profiles and the size of their smallest abductive explanations.
\end{abstract}

\end{frontmatter}

\section{Introduction}

Explaining the outcomes of artificial intelligence algorithms is one of the main challenges of present research in this field, with a plethora of competing approaches and venues to discuss them (see, e.g., the recent special issue edited by \citet{MilleretAlAIJ2022}).
This problem is traditionally less of an issue for voting rules, where social choice theorists have used the axiomatic approach to justify the use of a specific voting rule. More recently, a stream of papers have used such axiomatic properties to obtain justifications for the outcome of a voting rule on a given preference profile \citep{CaillouxEndrissAAMAS2016,peters2020explainable,NardiEtAlAAMAS2022}. 
Moreover, voting rules are no black boxes. They are usually given in the form of an explicit function or procedure that computes the winner of a given election. As already observed by \citet{ebadian2024explainable}, classical voting rules typically admit straightforward procedural explanations.

However, the realm of digital democracy can be led to implement arguably complex rules that require lengthy explanations to justify or explain how the outcome is computed from voters' preferences. To give two examples, the deliberation platform \emph{LiquidFeedback}\footnote{https://liquidfeedback.com/en/} uses the Schultze voting rule \citep{behrens2014principles}, and many participatory budgeting bodies are experimenting with the recently proposed method of equal shares\footnote{https://equalshares.net/} \citep{peters2021proportional}. 
Moreover, recent work is starting to consider social choice on large number of alternatives, be them a list of government proposals or sentences produced by LLMs in response to prompts (see, e.g., \citep{navarrete2024understanding,conitzer2024social,gudinoetal2024llm}). 
In this setting, even voting rules whose functioning is easy to explain---such as scoring rules---may profit from the development of magnifying lenses, in the form of well-studied and scientifically grounded algorithms for explanations, that are able to point at which subsets of the preference profile determines the top positions in the collective ranking, arguably augmenting the voters' trust in the outcome of the vote while at the same time providing the analyst with important information on the structure of the voters' preferences.

This paper argues that such tools can be obtained by suitably adapting notions from the growing body of research on formal explanations in machine learning, most notably \emph{abductive explanations} also called \emph{prime implicant explanations}~\citep{ijcai2018p708,darwiche2020reasons,marques-silva_logic-based_2022} and their dual \emph{contrastive explanations}~\citep{ignatiev_relating_2020}.
The underlying idea behind this endeavour is to explain the decision of supervised machine learning algorithms known as classifiers by identifying subsets of the feature space for which all possible completions do not change the classifier's decision (abductive explanations) or such that there exists a completion changing the decision (contrastive explanations, a special case of counterfactual explanations \citep{wachter2017counterfactual,liu2023unified}). 

We will focus on a specific class of voting rules known as scoring rules, which includes the plurality rule, $k$-approval rules, and the Borda rule.
Our choice is motivated by their computational properties---most notably the fact that computing a necessary winner can be done in polynomial time---as well as due to its wide application in recent digital democracy applications (see, e.g., \citep{lee2019webuildai,navarrete2024understanding}).


\paragraph{Our contribution.} We first adapt the definitions of abductive and contrastive explanations to the realm of voting rules in Section~\ref{sec:preliminaries}. 
We design an appropriate feature space on which explanations can be computed. Given our focus on scoring rules, we chose to represent profiles of preferences as \emph{rank matrices}, in which $n \times m$ non-binary features represent (in a non-anonymous way) which among $m$ candidates is placed in $k$-th entry by voter $i$ in its voting ballot. We also show a useful connection between formal explanations and the classical concepts of possible and necessary winner which will form the basis of our algorithms for computing explanations.

Computing formal explanations for binary classifiers is a search problem in an exponential space, but efficient algorithms have been proposed in the literature, often using SAT solvers. Computing formal explanations in voting requires tackling the additional problem of interconnected features, implied by the assumptions that preferences are complete, transitive, and irreflexive binary relations. We define and analyse our algorithms for the computation of formal explanations in voting in Section~\ref{sec:computation}.

Typically, the winner of a scoring rule on a preference profile can be explained by a large number of formal explanations. We are therefore interested in providing bounds on the size of the smallest ones.
Such size can be considered as an intrinsic measure of the richness or the complexity of a preference profile, and it has surprising links with the robustness of the resulting winner.
In Section~\ref{sec:size} we first prove a lower bound on the size of abductive explanations for the Borda rule, showing that it is linear in the number of voters. 
To complement our worst-case analysis, we conduct computer simulations using the recent tool of map of elections \citep{szufa2020drawing,ijcai2021p9}, showing an interesting negative correlation between the size of the smallest explanation and both the margin of victory of the winner and the intrinsic agreement of the preference profile.\footnote{The code and data of our experiment are available at \url{https://gitlab.irit.fr/ccontet1/axp-and-cxp-for-scoring-rules}}

\paragraph{Related work}


The closest work and the source of inspiration for our research is the recent stream of papers that justifies or explains the result of a voting rule with a logical calculus on sequences of axiomatic properties.
This approach was started by \citet{CaillouxEndrissAAMAS2016} for the Borda rule, then complemented by \citet{boixel2022calculus} with a full calculus. Algorithms for computing justifications were presented by \citet{NardiEtAlAAMAS2022} and implemented in a demo \citep{BoixelEtAlIJCAI2022}. \citet{peters2020explainable} gives minimal bounds on the length of the explanations. 
 
Our encoding of preference profiles in a space of non-independent features is an instance of classifiers under constraints (our constraints being the transitivity, completeness and asymmetry of users' preferences). Formal explanations under constraints are the subject of a recent paper by \citet{CooperAmgoudECAI2023}, from which we borrow the concept of irredundant explanations. 

Black-box machine learning techniques have shown to be effective in learning the behaviour of voting rules \citep{anil2021learning,burka2022voting}, but their potential for obtaining explanations is limited. A recent paper by \citet{KangHanXia2023} proposes the use of decision trees, producing human-readable diagrams of voting rules, albeit on small number of candidates. Explainability of voting rules is at the center of a related paper by \citet{ebadian2024explainable}, which explores the tradeoff between adding randomisation to voting rules while preserving their procedural explanatory power.

To the best of our knowledge this is the first paper to consider counterfactual reasoning to explain voting rules. However, abductive and constrastive explanation have direct connections with known and well-studied concepts in voting theory, such as necessary and possible winners \citep{konczak_voting_2005}, bribery and control \citep{faliszewski2016control}, defense against such attacks \citep{lu2020computational}, and communication complexity of voting rules \citep{ConitzerS05}. 
We explicit these connections
in our concluding section.


\section{Preliminaries}\label{sec:preliminaries}

In this section we introduce the notions of abductive and contrastive explanations and apply them to the realm of voting theory. 


\subsection{Formal explanations}

Formal explanations are tools recently developed for explainable artificial intelligence~\citep{ijcai2018p708,darwiche2020reasons,ignatiev_relating_2020,marques-silva_logic-based_2022,liu2023unified}.
Given a classification problem, the aim is to find inclusion-minimal subsets of features that are able to explain the classifier decision on a given instance.
For \emph{abductive explanations} we look for minimal subsets of features such that any extension does not change the classifier decision, and for \emph{contrastive explanations} we look for minimal subsets such that there exists an extension that reverts the classifier's decision.
Both types of explanations are based on the computation of minimal conjunctions of literals representing whether the value of a feature can vary.

As an explanatory example, assume that the decision of whether to allow a bank client to open a mortgage is done by a classifier on the basis of age and revenue. 
A negative decision by the classifier on an instance can be explained by an abductive explanation pointing out that the revenue is too low, so that the answer is negative irrespective of the age of the client. A contrastive explanation, on the other hand, would point out that by keeping the same age and changing the revenue it would be possible to obtain the mortgage.

Formally, following the notations used by \citet{marques-silva_logic-based_2022}, we are given a set of $N$ features $\mathcal{F}$, with each $i$-th feature having values in finite domain $\mathcal{D}_i$, and a set of classes $\mathcal{K}$. A feature space is defined by $\mathbb{F} = \mathcal{D}_1 \times \mathcal{D}_2 \times \dots \times \mathcal{D}_N$ and a classifier is any non-constant function $\kappa$ that maps the feature space $\mathbb{F}$ into the set of classes $\mathcal{K}$, i.e. $\kappa : \mathbb{F} \to \mathcal{K}$.

\begin{definition}\label{def:formalexp}
    Given $\mathrm{v} \in \mathbb{F}$, an abductive explanation (AXp) for the classification of $\mathrm{v}$ by $\kappa$ is any minimal subset  $\mathcal{X} \subseteq \mathcal{F}$ such that 
\begin{equation}\label{eq:axp}
\forall (\mathrm{x}\in \mathbb{F}).~ \left[\left(\bigwedge_{i\in\mathcal{X}}(x_i = v_i)\right) \implies (\kappa(\mathrm{x}) = \kappa(\mathrm{v}))\right]
\end{equation}

\noindent 
Given $\mathrm{v} \in \mathbb{F}$, a contrastive explanation (CXp) for the classification of $\mathrm{v}$ by $\kappa$ is any minimal subset $\mathcal{Y} \subseteq \mathcal{F}$ such that 
\begin{equation}\label{eq:cxp}
\exists (\mathrm{x}\in \mathbb{F}).~ \Bigg[\Bigg(\bigwedge_{i\in\mathcal{F}\setminus\mathcal{Y}}(x_i = v_i)\Bigg) \land (\kappa(\mathrm{x})\neq \kappa(\mathrm{v}))\Bigg]
\end{equation}
\end{definition}

Intuitively, given a specific instance, an AXp is a minimal subset of issue values to keep in order to ensure that the outcome of $\kappa$ remains unchanged, while a CXp is a minimal subset of issue values to erase in order to be able to change the outcome of $\kappa$.


\subsection{Elections and scoring rules}

An \emph{election} consists of a set $\agents$ of $\n$ agents, a set $\candidates$ of $\m$ candidates, and a \emph{preference profile} $\Profile = (\Profile_1,\dots ,\Profile_n)$, where each $\Profile_i$ is a linear order over $\candidates$, called a \emph{preference relation} or \emph{ballot}. We denote $\mathbb{P}$ the set of all possible profiles.
A \emph{non-resolute voting rule} $F$ is a function that maps $\mathbb{P}$ into a subset of candidates, i.e., $F:\mathbb{P} \to P(\candidates)$ with $P(\candidates)$ the power set of the set of candidates.

There exists a particular class of voting rules known as \emph{scoring rules}. Given a vector of weights $(w_1,w_2,\dots,w_\m)$, for each ballot the $i$\textsuperscript{th} candidate scores $w_i$ points. The winners are then the candidates with the highest total score over all the ballots. 
Formally, given a preference profile $\Profile$, let $\mathit{pos}(c,\Profile_i)$ be the position of candidate $c$ in the linear order $\Profile_i$ submitted by voter $i$, with the top-ranked candidate being in position 1. A scoring rule assigns to each candidate $c$ a score equals to $\sum_{i\in \agents} w_{\mathit{pos}(c,\Profile_i)}$, and the candidates with the highest score are declared the winners.
Notable examples include the Borda rule defined by scoring vector $(w_1,w_2,\dots,w_\m) = (\m-1, \m-2,\dots,0)$, and $k$-approval rules defined  by vectors $(1,\dots,1,0\dots,0)$, with $k$ being the number of 1s in the scoring vector (with 1-approval being the plurality rule).
For an introduction on voting rules we refer to \citet{BRAMS2002173} and \citet{ZwickerHandbook16}.

\begin{example}\label{ex:first}
Consider four candidates $A, B, C, D$ and four voters. We represent the preferences of the voters with the rows of the following matrix, with the most preferred candidate on the left and the least preferred on the right:
$$\begin{tabular}{ c c }
$\R = \begin{bmatrix}
    A & B & C & D\\
    B & C & D & A\\
    A & D & C & B\\
    D & C & A & B
\end{bmatrix}$ & 
\begin{tabular}{c}
    voter 1\\
    voter 2\\
    voter 3\\
    voter 4\\
\end{tabular}
\end{tabular}
$$

The Borda scores of candidate $A$ is 7, since it is ranked first by two voters and third by one voters. For the remaining candidates B, C, and D, the Borda scores are 5, 6, and 6, respectively. Hence, $A$ is the winner for the Borda rule on this preference profile.
\end{example}

\subsection{Formal explanations for elections}

Voting rules can be seen as special cases of classifiers that take profiles of linear orders as input and output a set of winning candidates. 
However, this representation depends on how input profiles are encoded into a feature space. This choice is crucial to formal explanations, as the encoding into a feature space determines the space of possible explanations (see Definition~\ref{def:formalexp}), with important consequences on their expressiveness and computational complexity.

In the literature on (computational) social choice, incomplete preferences are usually represented as partial orders (see, e.g., \citep{konczak_voting_2005,xia_determining_2011}). 
However, for the specific case of scoring rules we propose the use of \emph{partial rank matrices} in view of their more compact representation.

\begin{definition}
    A \textbf{rank matrix} $\R = (\R_1,\dots ,\R_\n)$ is an $\n \times \m$ matrix, where each row $\R_i$ is a permutation of the set of candidates $\candidates$.
\end{definition}

Profiles can naturally be represented with rank matrices (see Example~\ref{ex:first}). In this context, each row represents a voter's ballot as a linear order over candidates. Given a candidate $c \in \candidates$ and $k\leq \m$,  $\R_{i,k} = c$ means that $c$ is in the $k$\textsuperscript{th} position in the ballot of voter~$i$.
Throughout this paper, we will use interchangeably the terms vote profile and rank matrix.

\begin{definition}
    A \textbf{partial rank matrix} $\R = (\R_1,\dots ,\R_\n)$ is an $\n \times \m$ matrix
    with values in $\candidates\cup \{\nulle\}$, such that on every row $\mathcal R_i$ each element of $\candidates$ appears at most once.
\end{definition}

If $\R_{i,k} = \nulle$, we say that $k$ is a \emph{free entry} in $\R_i$.  Conversely, $k$ is a \emph{locked entry} in $\R_i$ if $\R_{i,k} \neq \nulle$. When displayed in our examples, free entries will be represented by middots ($\cdot$). We define the size of a partial rank matrix $\R$ (respectively ballot $ \R_i$), noted $|\R|$ (respectively $ |\R_i|$), as the number of its non-null entries.

\begin{definition}
    Given two partial rank matrices $\R$ and $\R'$, we say that $\R'$ is an \textbf{extension} of $\R$, denoted $\R \subseteq \R'$, if for all $(i,k)$ such that $\R_{i,k} \neq \nulle$ we have that $\R_{i,k} = \R'_{i,k}$. 
    We denote with $Ext(\R)$ the set of all \textbf{complete extensions} of a partial rank matrix $\R$, i.e., extensions of $ \R $ with no $\nulle$ value.
\end{definition}

Given two rank matrices $\R'$ and $\R''$ such that $\R' \subseteq \R''$, we define $\R = \R''\setminus{\R'}$ to be the partial rank matrix such that $\R_{i,k} = \nulle$ for all $(i,k)$ such that $\R'_{i,k} \neq \nulle$, and $\R_{i,k} = \R''_{i,k}$ for all other entries.
We can now adapt Definition~\ref{def:formalexp}
for scoring rules. 

\begin{definition}\label{def:votingexp}
Given a complete rank matrix $\R$, a voting rule $F$ and a winning candidate $w \in F(\R)$, 
an AXp of $w \in F(\R)$ is a $\subseteq$-minimal partial rank matrix $\mathcal{X}$ such that $\mathcal{X} \subseteq \R$ and
\begin{equation}\label{eq:axpvr}
\forall (\R' \in \mathbb{P}) \text{, } \mathcal{X} \subseteq \R' \implies w \in F(\R').
\end{equation}
A CXp of $w \in F(\R)$ is a $\subseteq$-minimal partial rank matrix $\mathcal{Y}$  such that $\mathcal{Y} \subseteq \R$ and
\begin{equation}\label{eq:cxpvr}
\exists (\R' \in \mathbb{P})\text{, } (\R\setminus{\mathcal Y}) \subseteq \R' \,\text{ and }\, w \not \in F(\R').
\end{equation}
\end{definition}

\begin{example}\label{ex:AXp}
Consider the following partial rank matrix:
$$\X^1 = \begin{bmatrix}
    A & B & \cdot & \cdot \\
    \cdot & C & D & \cdot\\
    A & D & \cdot & \cdot\\
    \cdot & \cdot & \cdot & B
\end{bmatrix}$$

\noindent
In any complete extension of $ \X^1 $, candidate $A$ scores 3 points under the Borda rule with the ballot of voter 1, 3 points with voter 3, and at least 1 point with voter 4 (since for that voter, $ B $ is already in the last position); so the overall score of $ A $ cannot be less than 7. Conversely, $ B $ may score $ 3 $ with voter 2, but cannot score more than 1 with voter 3 (in the second-to-last position), so the overall Borda score for B cannot be more than 6. 
Thus $ A $ beats $ B $ with the Borda rule in any complete extension of $\X^1$, and it is not difficult to check that neither $ C $ nor $ D $ can beat $ A $; although there can be ties, $ A $ is assured to be one of the Borda winners.
Moreover, if any of the non-null entries in $ \X^1 $ is freed (i.e., replaced with $ \cdot $), then it can be shown that there is a complete extension in which $ A $ is not a Borda winner anymore. 
%
Since $ \X^1 \subseteq \R $, this shows that $ \X^1 $ is an AXp for $ A \in \Borda(\R) $.
\end{example}

\begin{example}\label{ex:CXp}
Consider 
$ \Y^1 $ below and its complement wrt. $ \R $:
$$
  \Y^1 = \begin{bmatrix}
    A & \cdot & C & \cdot \\
    \cdot & \cdot & \cdot & \cdot\\
    \cdot & \cdot & \cdot & \cdot\\
    \cdot & \cdot & \cdot & \cdot
\end{bmatrix}
\qquad
\R \setminus \Y^1 =\begin{bmatrix}
    \cdot & B & \cdot & D\\
    B & C & D & A\\
    A & D & C & B\\
    D & C & A & B
\end{bmatrix}
$$
If $ \R^2 $ is the completion of $ \R \setminus \Y^1 $ with $ C $ in the first position for voter 1, and $ A $ in the third position, then the Borda score of $ A $ in $ \R^2 $ is $ 1+0+3+1 = 5 $, the score of $ C $ is $ 3+2+1+2 = 8 $, thus $ A \notin \Borda(\R^2) $. This shows that $ \Y^1 $ is a \CXp\ for $ A \in \Borda(\R) $.
\end{example}


As observed previously, the feature space we just defined is not composed of independent features (the cells of a rank matrix). The following example calls for a refinement of the notion of AXp.

\begin{example}
Consider the two partial rank matrices $ \X^2 $ and $ \X^3 $ below, of which the rank matrix $ \R $ of Example~\ref{ex:first} is a complete extension. Observe that $\X^1$ and $\X^2$ only differ in the first row:
$$\X^2 = \begin{bmatrix}
    \cdot & B & C & D\\
    B & \cdot & \cdot & \cdot\\
    A & D & \cdot & \cdot\\
    \cdot & \cdot & \cdot & B
\end{bmatrix}
\qquad
\X^3 = \begin{bmatrix}
    A & \cdot & C & D\\
    B & \cdot & \cdot & \cdot\\
    A & D & \cdot & \cdot\\
    \cdot & \cdot & \cdot & B
\end{bmatrix}
$$
Both $ \X^2 $ and $ \X^3 $ are AXps for $ A \in \Borda(\R) $. However, it can be argued that they represent the same explanation, because the first row of $ \X^2 $ and $ \X^3 $ have a single possible extension, namely $ [A \; B \; C \; D] $. In other words: $ \X^2 $ and $ \X^3 $ represent the same explanation, which is better represented with the full ballot $ [A \; B \; C \; D] $ on the first row. 
\end{example}

As the example above shows, we can have multiple AXps (up to $m$) which differ only by one null entry in a row, and this is due to the constraints defining a linear order of candidates. To avoid the redundancy caused by multiple equivalent AXps we propose the following:

\begin{definition}
Given a complete rank matrix $\R$, a voting rule $F$ and a winning candidate $w \in F(\R)$, an \textbf{irredundant abductive explanation}, or \newAXp, of $w \in F(\R)$ is a $\subseteq$-minimal partial rank matrix $\mathcal{X}$ such that $\mathcal{X} \subseteq \R$, it verifies equation~(\ref{eq:axpvr}), and such that every row has either none or at least 2 $null$ entries.
\end{definition}


Formal explanations under constraints have been studied by \citet{CooperAmgoudECAI2023}. Our \emph{\newAXp}s correspond to their notion of \emph{subset-minimal, coverage-based prime-implicant explanation (mCPI-Xp)}. 
Note that an AXp typically is also an \emph{iAXp}, except when it has one or more rows with one null entry only. Given their closeness, in the remainder of the paper we will talk about AXps\ in general, except when discussing algorithms for their computation.

\subsection{Necessary winner and explanations}\label{ssec:NW}

The well-studied concept of necessary and possible winner of voting rules~\citep{konczak_voting_2005} can be used to obtain a useful equivalent formulation for Definition~\ref{def:votingexp}. 
Recall that given a candidate $c \in \candidates$, a voting rule $F$, and a partial rank matrix $\R$, $c$ is a necessary winner if for every extension $\R'$ of $\R$ we have that $c \in F (\R')$, and we write $c \in \NW{F}{\R}$. Similarly, $c$ is a possible winner for $\R$ and $F$ if there exists an extension $\R'$ of $\R$ such that $c \in F(\R')$, and we write $c \in \PW{F}{\R}$. It is now straightforward to show the following: 

\begin{proposition}\label{prop:xpnec}
Given a complete rank matrix $\R$, a voting rule $F$ and a winning candidate $w \in F(\R)$,
$\mathcal{X}$ is an AXp of $\R$ iff $\mathcal{X}$ is a $ \subseteq $-minimal partial rank matrix s.t. $\mathcal{X} \subseteq \R$ and
$w \in \NW{F}{\mathcal{X}}$.
Moreover, $\mathcal{Y}$ is a CXp of $\R$ iff $\mathcal{Y}$ is a $ \subseteq $-minimal partial rank matrix s.t. $\mathcal{Y} \subseteq \R$ and
$w \not \in \NW{F}{\R\setminus{\mathcal{Y}}}$.
\end{proposition}

For scoring rules, the problem of deciding whether or not a candidate is a necessary winner can efficiently be solved thanks to a characterization based on candidates' minimal and maximal achievable scores~\citep{konczak_voting_2005}.
Given its importance for the rest of the paper, we reframe the original result here with our notation.

Given a (possibly partial) rank matrix $\R = (\R_1,\dots ,\R_\n)$, let us denote the score obtained by $c$ in $\R$ as $\sigma_{\R}(c)$ and the score in a single ballot $\R_i$ as $\sigma_{\R_i}(c)$. Clearly, 
$\sigma_{\R}(c) = \sum_{\R_i \in \R} \sigma_{\R_i}(c)$.
For any partial rank matrix $\R$ and any candidate $c$, we introduce $\sigma^{min}_{\R}(c) = min_{\R' \in Ext(\R)} \sigma_{\R'}(c)$ (resp. $\sigma^{max}_{\R}(c) = max_{\R' \in Ext(\R)} \sigma_{\R'}(c)$) to be the minimal (resp. maximal) score that candidate $c$ can obtain in any complete extension of $\R$. \citet{konczak_voting_2005} proved the following characterization of necessary winners: 

\begin{proposition}\label{prop:charaNW}
    Let $F$ be a scoring rule, $\R$ a (possibly partial) rank matrix, and $c$ a candidate. $c \in \NW{F}{\R}$ if and only if for all $c' \in \candidates\setminus{\{c\}}$, 
    \begin{equation}
        \sigma^{min}_{\R}(c) \geq \sigma^{max}_{\R}(c')
    \end{equation}
\end{proposition}

\forreviewers{\noindent A more detailed analysis and proof of this proposition can be found in the extended version of this paper~\citep{extended}.}

\begin{toappendix}
\begin{proof}[Proof of proposition~\ref{prop:charaNW}]
In the original paper presenting the concept of necessary winner~\citep{konczak_voting_2005}, the authors use a different representation of partial profiles. Where we use partial rank matrices, they use partial orders on candidates.

This does not impact the characterisation of necessary winner which is independent of the representation. Indeed, for a profile $\Profile$, a voting rule $F$ and a winning candidate $c\in F(\Profile)$, if for all candidate $c'$ different from $c$ and all $\Profile'$ a such that $\Profile$ is an extension of $\Profile'$, we have that $\sigma^{min}_{\Profile'}(c) \geq \sigma^{max}_{\Profile'}(c')$, then $\sigma_{\Profile'}(c) \geq \sigma_{\Profile'}(c')$ because by definition we have $\sigma_{\Profile'}(c) \geq \sigma^{min}_{\Profile'}(c)$ and $\sigma^{max}_{\Profile'}(c') \geq \sigma_{\Profile'}(c')$.

\citet{konczak_voting_2005} present this result as an equivalence. However as pointed out by \citet{xia_determining_2011} this is a mistake as the following counterexample shows. Consider four candidates $A$, $B$, $C$, $D$, three voters with the Borda rule. The first vote is the partial order $A \succ B$, $A \succ C$, $A \succ D$, $B \succ C$, $B \succ D$. The second vote is $A \succ C$, $A \succ D$, $B \succ A$, $B \succ C$, $B \succ D$. The third is $A \succ B$. $A$ is in first position in vote 1, in second in vote 2 and cannot be lower than third in vote 3. Hence, $\sigma^{min}(A)=6$. Similarly, $\sigma^{max}(B) = 7$, $\sigma^{max}(C) = 5$ and $\sigma^{max}(D) = 5$. $A$ ties with $B$ in the first two votes but beats $B$ in the third vote so $A$ is ranked above $B$. Since $\sigma^{min}(A) > \sigma^{max}(C)$ and $\sigma^{min}(A) > \sigma^{max}(D)$, $A$ is ranked above $C$ and $D$. Thus, we know that $A$ will always be first and hence is a necessary winner. However, we have $\sigma^{min}(A) < \sigma^{max}(B)$.

Even if the equivalence does not hold in the original setting of partial orders, we now show that it holds when preference profiles are represented as rank matrices. Indeed, in our context, minimizing the score of a necessary winner $w$ and maximizing the score of any other candidate $c$ are two independent goals. 
Hence, given a partial rank matrix $\R$, for every candidate $c$, there exist a complete rank matrix $\R_c$ such that $\R \subseteq \R_c$, $\sigma_{\R_c}(w) = \sigma^{min}_{\R}(w)$ and $\sigma_{\R_c}(c) = \sigma^{max}_{\R}(c)$. Since $w$ is a necessary winner of partial rank matrix $\R$, for any extension $\R'$ of $\R$, we have $\sigma_{\R'}(w) \geq \sigma_{\R'}(c)$. Applying this to each $\R_c$, we have $\forall c \in \candidates,\,\sigma^{min}_{\R}(w) \geq \sigma^{max}_{\R}(c)$.
To construct such extension of $\R$, we work ballot by ballot. If at least $w$ or $c$ is locked achieving both goals simultaneously then, is trivial. If both $w$ and $c$ are free, there are at least two free entries in the ballot. Hence, we can put $c$ in the most preferred free entry and $w$ in the least preferred free one without any problem.
\end{proof}
\end{toappendix}

\begin{example}\label{ex:NW}
Consider the partial rank matrix $\X^1 $ from Example~\ref{ex:AXp}. The table below gives the minimal and maximal Borda scores $ \sigma^{min}_{\X^1}(A) $ and $ \sigma^{max}_{\X^1}(B) $
as well as $ \sigma^{max}_{\X^1}(C) $ and  $ \sigma^{max}_{\X^1}(D) $:

$$
\X^1 = \begin{bmatrix}
    A & B & \cdot & \cdot \\
    \cdot & C & D & \cdot\\
    A & D & \cdot & \cdot\\
    \cdot & \cdot & \cdot & B
\end{bmatrix}\qquad 
\arraycolsep2pt
\begin{array}{ |c|c|c|c|c| } 
 \hline
  & A & B & C & D \\
  \hline
 \multirow{2}{*}{$\sigma^{min|max}_{\X^1}$}  & min & max & max & max \\ 
  & 7 & 6 & 7 & 7 \\
 \hline
\end{array}
$$

\noindent
Using Proposition~\ref{prop:charaNW} we can infer that $A \in \NW{\Borda}{R^1}$. This, combined with the fact that freeing any other non-null entry in $ \X^1 $ would decrease $ \sigma^{min}_{\X^1}(A) $, or increase $ \sigma^{max}_{\X^1} $ for some of the other candidates, shows that $ \X^1 $ is an AXp of $A\in \Borda(\R)$ by Proposition~\ref{prop:xpnec}.

Similarly, consider the partial rank matrix $ \Y^1 $ of Example~\ref{ex:CXp}. The table below gives 
minimal and maximal Borda scores for $ A, B $, $ C $ and $ D $ in any complete extension of $ \R \setminus \Y^1 $:
$$
\R\setminus\Y^1 \!=\! \begin{bmatrix}
    \cdot & B & \cdot & D\\
    B & C & D & A\\
    A & D & C & B\\
    D & C & A & B
\end{bmatrix}
\;\;
\arraycolsep1pt
\begin{array}{ |c|c|c|c|c| } 
 \hline
  & A & B & C & D \\
  \hline
\multirow{2}{*}{$\sigma^{min|max}_{\R\setminus\Y^1}$}  & min & max & max & max \\  
  & 5 & 5 & 8 & 6 \\
 \hline
\end{array}
$$

\noindent
Observe that $ \sigma^{min}_{\R\setminus\Y^1}(A) < \sigma^{max}_{\R\setminus\Y^1}(C) $, thus $ A \notin \NW{\Borda}{\R\setminus\Y^1} $.
Moreover $ \Y^1 $ is a minimal partial rank matrix with this property, making $ \Y^1 $ a \CXp\ for $ A \in \Borda(\R) $ by Proposition~\ref{prop:xpnec}.

\end{example}


\section{Computing explanations for voting rules}\label{sec:computation}

Existing algorithms to compute formal explanations range from finding one AXp or CXp, obtaining one (cardinal-wise) smallest AXp or CXp, and enumerating all AXps and CXps (we refer to \citep{marques-silva_logic-based_2022} for a recent survey). All these algorithms are specializations of algorithms used to compute Minimal Unsatisfiable Sets (MUS) and Minimal Correction Sets (MCS)~\citep{ignatiev_relating_2020}. 
In this section we adapt these algorithms to the computation of formal explanations for scoring rules.

As previously observed, in our settings the features are not independent, since they are entries of a rank matrix. 
Hence, our algorithms have to include additional guardrails to 
ignore redundant explanations generated by the non-independence of our features.



\subsection{Computing one explanation}

Algorithm~\ref{alg:findCXp} finds one CXp for a given rank matrix and one of the winning candidates. 
Because the non-independence of features does not play a role in the computation of CXps, it is a straightforward adaptation of the classical algorithm for finding contrastive explanations for arbitrary classifiers. 
Indeed, when computing CXps we look for the minimal amount of matrix entries to remove to be able to change the winner, and removing a single entry in a row of a rank matrix does not change the set of possible extensions.
Algorithm~\ref{alg:findCXp} goes across each entry of the rank matrix and locks as many as possible while keeping the constraint $w \in \NW{F}{\R\setminus{\Y}}$. If this condition is broken, the algorithm rolls back by one step and frees the entry that has just been locked. 
Since it is easier to verify that $w \in \NW{F}{\Yb}$ where $\Yb = \R\setminus{\Y}$, the algorithm works on the complement of the CXp, namely $\Yb$.
Due to the monotonicity of explanations (i.e., if we have enough information to declare the winner, having even more information will also be enough to declare the winner), the entry removed and then added back will not have to be tested again later.

Algorithm~\ref{alg:findCXp} and, later, Algorithm~\ref{alg:findiAXp}, take a seed in input parameters. This seed will be used for computations in Algorithms~\ref{alg:enumXp} and~\ref{alg:findSiAXp} to have some control on the explanation returned. 
The seed plays the role of a shortcut: one can view it as if all the entries it contains were the first iterations of the for loop on line 2. Since we have $w \not \in \NW{F}{\R\setminus{\Se}}$, we know that no rollback would have occurred.
Algorithms~\ref{alg:findCXp} and~\ref{alg:findiAXp} can be run with the whole rank matrix as the seed to output a minimal explanation.

\begin{algorithm}[h]
\caption{\textsc{findCXp} \--- Finding one CXp}\label{alg:findCXp}
\KwData{Rank matrix $\R$, scoring rule $F$, winning candidate $w \in F(\R)$, partial rank matrix $\Se$ as a seed s.t. $\Se\subseteq\R$ and $w \not \in \NW{F}{\R\setminus{\Se}}$}
\KwResult{a CXp}
$\Yb \gets \R\setminus{\Se}$ \tcp*{$\Y \gets \Se$}
\For{$(i,j) \in [|1,\n|]\times[|1,\m|] \text{ and } \Se_{i,j} \neq null$}{
    $\Yb_{i,j} \gets \Se_{i,j}$ \tcp*{$\Y_{i,j} \gets null$}
	\If(\tcp*[f]{if $w \in \NW{F}{\R\setminus{\Y}}$}){$w \in \NW{F}{\Yb}$ }{
        $\Yb_{i,j} \gets null$\tcp*{$\Y_{i,j} \gets \Se_{i,j}$}
	}
}
\Return{$\R\setminus{\Yb}$} \tcp*{return $\Y$}
\end{algorithm}

Algorithm~\ref{alg:findiAXp} finds one irredundant AXp for a given rank matrix and one of the winning candidates. Because of the duality between the definitions of AXp and CXp (see, e.g., \citep{ignatiev_relating_2020}), its structure is very similar to Algorithm~\ref{alg:findCXp} with additional steps to take into account the non-independence of features.
This takes the form of both an additional constraint on the seed (namely that $\forall i$, $|\Se_i| \neq \m - 1$) and a sub-function \textsc{ensureIrr} described in Algorithm~\ref{alg:ensureIrr}, that makes sure that no row in the resulting explanation has one only free entry. This mechanism triggers when, for the first time, an entry is freed in a row (line 4 of Algorithm~\ref{alg:findiAXp}). In that case, Algorithm~\ref{alg:ensureIrr} looks for another entry to free in the same ballot. If one is found the execution goes on normally (break instruction on line 6) and if none, the entry initially freed belongs to the AXp (line 7 after the main for loop). 

\begin{algorithm}[t]
\caption{\textsc{findiAXp} \--- Finding one \newAXp}\label{alg:findiAXp}
\KwData{Rank matrix $\R$, scoring rule $F$, winning candidate $w \in F(\R)$, partial rank matrix $\Se$ as a seed s.t. $\Se\subseteq\R$, $w \in \NW{F}{\Se}$ and $\forall i$, $|\Se_i| \neq \m - 1$}
\KwResult{an \newAXp}
$\X \gets \Se$\;
\For{$(i,j) \in [|1,\n|]\times[|1,\m|] \text{ and } \Se_{i,j} \neq null$}{
	$\X_{i,j} \gets null$\;
    \eIf{$|\X_i| = \m - 1$}{
        \textsc{ensureIrr}($\R,F,w,\Se,\X,i,j$)
    }{
    	\If{$w \not \in \NW{F}{\X}$}{
        	$\X_{i,j} \gets \Se_{i,j}$\;
    	}
    }
}
\Return{$\X$}
\end{algorithm}

\begin{algorithm}[t]
\caption{\textsc{ensureIrr} \--- Ensuring irredundancy of the iAXp currently computed}\label{alg:ensureIrr}
\KwData{Rank matrix $\R$, scoring rule $F$, winning candidate $w \in F(\R)$, partial rank matrix $\Se$, partial rank matrix $\X$ s.t. $\X\subseteq\Se$, $w \in \NW{F}{\X}$, index $i$ s.t. $|\X_i| = \m - 1$, index $j$ s.t. $\X_{i,j} = null$}
\For{$ j_2 \in [|1,\m|] \text{ and } j_2 \neq j$}{
    $\X_{i,j_2} \gets null$\;
    \eIf{$w \not \in \NW{F}{\X}$}{
        $\X_{i,j_2} \gets \Se_{i,j_2}$\;
    }{
        \textbf{break}\;
    }
}
$\X_{i,j} \gets \Se_{i,j}$\;
\end{algorithm}



\begin{proposition}
    Finding one CXp (Algorithm~\ref{alg:findCXp}) can be done in $\Theta(\n\m^2)$ and finding one \newAXp ~(Algorithm~\ref{alg:findiAXp}) in $\Theta(\n\m^3)$.
\end{proposition}

\begin{proof}[Proof sketch]
Algorithm~\ref{alg:findCXp} goes across all the $\n\m$ entries of the rank matrix, tries to remove it and checks if it changes the outcome of the necessary winner test. Algorithm~\ref{alg:findiAXp} works similarly but for the addition of Algorithm~\ref{alg:ensureIrr}, which goes through a whole row adding a complexity factor of $\m$ in complexity.
Hence, the complexity of finding explanations greatly depends on our ability to decide whether or not a candidate is a necessary winner of a partial rank matrix. To do this, we use the characterisation introduced in Proposition~\ref{prop:charaNW}.
Since rows of a rank matrix are independent from each other, the minimal (resp. maximal) score can be decomposed as the sum of minimal (resp. maximal) scores on each row: $\sigma^{min}_{\R}(c) = \sum_{\R_i \in \R} \sigma^{min}_{\R_i}(c)$ (resp. $\sigma^{max}_{\R}(c) = \sum_{\R_i \in \R} \sigma^{max}_{\R_i}(c)$). It is possible to compute minimal and maximal possible scores of all candidates by simply scanning through the whole row of length $\m$. By repeating it for the $\n$ rows, we have a straightforward algorithm to check if a candidate is a necessary winner which runs in $\Theta(\n\m)$.
However, in our setting, it is possible to improve this bound. 
Observe that the necessary winner check in our algorithms is done repeatedly on rank matrices differing by only one entry. 
We can therefore keep in memory the minimal and maximal possible scores for all rows, and at each step of the for loop simply update the score of the row that has been modified, which as discussed above can be done in $\Theta(\m)$.
\end{proof}


\subsection{Enumeration and smallest explanations}


Algorithm~\ref{alg:enumXp} enumerates all iAXps and CXps by exploring the whole search space. It iteratively generates a seed which will lead to a new CXp or iAXp with a call to an NP oracle. In this case, the problem of finding new instances to explore is encoded in a SAT formula. Every entry of the rank matrix is associated to a literal where a \textit{True} assignment means that the entry is fixed. 
A SAT solver is then used to efficiently find new instances  thanks to the monotonicity of explanations. 
An upper bound on the total number of AXps and CXps is $2\binom{\n\m}{\left \lfloor\frac{\n\m}{2}\right \rfloor}$, which can be derived via Sperner's Theorem (for a deeper analysis of this mechanism see~\citep{liffiton_fast_2016}).
Finally, to prevent issues with the non-independence of our features in Algorithm~\ref{alg:enumXp} and later Algorithm~\ref{alg:findSiAXp}, instances containing a row with only one free entry are removed (lines 2 to 3). The added clauses can be read as: for each entry $i,j_0$, if $x_{i,j_0}$ is false (the entry is free), then $\bigvee_{j \neq j_0} \neg x_{i,j}$ is true (at least another entry in the row is free).


\begin{algorithm}[h]
\caption{\textsc{enumXp} \--- Enumerating all \newAXp s and CXps}\label{alg:enumXp}
\KwData{Rank matrix $\R$, scoring rule $F$, winning candidate $w \in F(\R)$}
\KwResult{List of all \newAXp s and CXps}
$\Map \gets \text{True}$\;
\For{$(i,j_0) \in [|1,\n|]\times[|1,\m|]$}{
    $\Map \gets \Map \land \left(x_{i,j_0} \lor \bigvee_{j \neq j_0} \neg x_{i,j}\right)$\;
}
\While{\Map~is satisfiable}{
    $u \gets \textsc{Sat}(\Map)$\;
    $\Se \gets \{\R_{i,j} \text{ if } u_{i,j} \text{ else } null \text{; } (i,j) \in [|1,\n|]\times[|1,\m|]\}$\;
    \eIf{$w \in \NW{F}{\Se}$}{
        $\iAXp \gets \textsc{findiAXp}(\R,F,w,\Se)$\;
        $\iAXps \gets \iAXps \cup \iAXp$\;
        $\Map \gets \Map \land \left(\bigvee_{(i,j) \in \iAXp} \neg x_{i,j}\right)$\;
    }{
        $\CXp \gets \textsc{findCXp}(\R,F,w,\Se)$\;
        $\CXps \gets \CXps \cup \CXp$\;
        $\Map \gets \Map \land \left(\bigvee_{(i,j) \in \CXp} x_{i,j}\right)$\;
    }
}
\Return{(\iAXps,\CXps)}
\end{algorithm}

When selecting one of the many possible formal explanations we will be interested in selecting the smallest one.
Recall that the size of a partial rank matrix is the number of its non-null entries. We can therefore define the set of smallest iAXps, denoted as \shortAXp, as the $\argmin_{\{\mathcal X \mid \mathcal X \text{ is \newAXp ~of } w\in F(\R)\}} |X|$.
Finding a smallest CXp is relatively easy and can be done by computing a smallest cost solution to a SAT problem. 
However, finding a \shortAXp ~is generally harder. Algorithm~\ref{alg:findSiAXp} is based on a previous solution which uses the hitting set duality of MUS and MCS and iteratively computes minimum (cardinal-wise) solutions to a hitting set problem \citep{10.1007/978-3-319-23219-5_13}. 

\begin{algorithm}[h]
\caption{\textsc{findSiAXp} \--- Finding one smallest iAXp}\label{alg:findSiAXp}
\KwData{Rank matrix $\R$, scoring rule $F$, winning candidate $w \in F(\R)$}
\KwResult{a smallest iAXp}
$\Map \gets \text{True}$\;
\For{$(i,j_0) \in [|1,\n|]\times[|1,\m|]$}{
    $\Map \gets \Map \land \left(x_{i,j_0} \lor \bigvee_{j \neq j_0} \neg x_{i,j}\right)$\;
}
\While{true}{
    $u \gets \textsc{MinimumHittingSet}(\Map)$\;
    $\Se \gets \{\R_{i,j} \text{ if } u_{i,j} \text{ else } null \text{; } (i,j) \in [|1,\n|]\times[|1,\m|]\}$\;
    \eIf{$w \in \NW{F}{\Se}$}{
        \Return{$\Se$}
    }{
        $\CXp \gets \textsc{findCXp}(\R,F,w,\Se)$\;
        $\Map \gets \Map \land \left(\bigvee_{(i,j) \in CXp} x_{i,j}\right)$\;
    }
}
\end{algorithm}


\section{Smallest abductive explanations for Borda}\label{sec:size}

One of the main challenges of dealing with formal explanations is the existence of a large number of possible explanations. 
Indeed, the simple rank matrix introduced in Example~\ref{ex:first} already has 14 different iAXps and 17 CXps.
We also run preliminary experiments on a simplified version of the preference profile used by~\citet{peters2020explainable} as their main example, with 8 voters and 4 candidates, and we obtained 3244 iAXps and 321 CXps. 
As iAXps and CXps can be used to improve trust and understanding of a voting rule, the size of the \emph{smallest} iAXps is of primary importance. In this section we provide tight lower bounds on sAXps complemented by experimental results using a suitably defined map of elections. 
Results in this section are restricted to the study of smallest AXps for the Borda rule. Naturally, they also apply to smallest iAXps.



\subsection{The size of abductive explanations for Borda}

We first give a lower bound on the size of abductive explanations for the Borda rule.
The proof is non-trivial, and uses a suitably defined notion of normal form for ballots of rank matrices, hinging on the characterisation of necessary winners proved in Proposition~\ref{prop:charaNW}.
\forreviewers{We give a sketch of the proof here, a detailed proof can be found in the extended version of this paper~\citep{extended}.}


\begin{theoremrep}\label{thm:lowerbound}
    Let $\R$ be a rank matrix with $\n$ voters and $\m$ candidates s.t. $w \in \candidates$ is a Borda winner of $\R$.
    For all AXp $\mathcal{X}$ of $\R$, we have:
    \begin{equation}\label{eq:lowerbound}
        |\mathcal{X}| \geq \n - \left \lfloor \frac{\n}{\m} \right \rfloor
    \end{equation}
\end{theoremrep}

\begin{proofsketch}
First, we observe that to bound the size of AXps of an arbitrary rank matrix, we can construct a different rank matrix which admits an AXp of size smaller or equal to the original matrix. This observation is at the basis of our normal form construction below.
Second, let a \emph{weak AXp} be any partial rank matrix verifying constraint (\ref{eq:axp}) but that is not necessarily minimal. Because of subset-minimality, smallest weak AXps are AXps, and since AXps are also weak AXps, then proving (\ref{eq:lowerbound}) restricted to weak AXps is equivalent to proving the bound in Theorem~\ref{thm:lowerbound}.

Our proof then uses the characterisation proved in Proposition~\ref{prop:charaNW}: if $ \X $ is an AXp for $ w \in \Borda(\R) $, $ w $ is a necessary winner of $ \X $ if and only if $ \sigma^{min}_{\X}(w) \geq \sigma^{max}_{\X}(c') $ for every other candidate $ c' $. 
We now define $ \Delta^ w_{\X}(c') = \sigma^{min}_{\X}(w) - \sigma^{max}_{\X}(c') $, and we call $ \Delta^w_{\X} = \sum_{c'\neq c} \Delta^w_{\X}(c') $ the total margin of victory. 
Since $ w $ is a necessary winner for $ \X $, then $ \Delta^w_{\X} \geq 0 $. 
We also show that the computation of the total margin of victory can be decomposed as the sum of \emph{total score margins}, computed ballot by ballot. Hence, in the start of the proof we will work at the level of a single ballot.

The main steps of the proof continues as follows, starting from any AXp $\mathcal X$ of an arbitrary rank matrix:
\begin{enumerate}
\item We prove that there exists a ballot $ \X_i' $ (possibly part of a different weak AXp) which has a particular form, which we call ``normal form'', such that $ | \X_i' | = | \X_i | $ and $ \Delta^w_{\X_i'} \geq \Delta^w_{\X_i} $. This normal form is obtained by repeated permutations of candidates, to bring (or insert) $ w $ at the top of the ballot, and ``descending''  some other candidates as much as possible while not decreasing $ \Delta^w $.
\item Thanks to our definition of normal form, we are able to prove that $ \Delta^w_{\X_i'} \leq (m-1)(m|\X_i'| - n(m-1)) $.
\item Summing on voters we have $ \Delta^w_{\X'} \leq (m-1)(m|\X'| - n(m-1)) $, which implies that $ \Delta^w_{\X} \leq (m-1)(m|\X| - n(m-1)) $.
\item Recall that $ 0 \leq  \Delta^w_{\X} $ since $w$ is a necessary winner, thus we have $0 \leq (m-1)(m|\X| - n(m-1))$ which implies the result.
\end{enumerate}
\vspace{-6mm}
\end{proofsketch}

\begin{toappendix}
Building on notations introduced in Section~\ref{ssec:NW}, for a given partial rank matrix $\R = (\R_1,\dots ,\R_\n)$, we define the score margin of $c$ w.r.t. $c'$ for $\R_i$ as $\smo{\R_i}{c}{c'} = \sigma^{min}_{\R_i}(c) - \sigma^{max}_{\R_i}(c')$ and the margin of victory of $c$ w.r.t. $c'$ for $R$ as $\smo{\R}{c}{c'} = \sum_{\R_i \in \R} \smo{\R_i}{c}{c'}$. The inequality of Proposition~\ref{prop:charaNW} can now be rewritten as $\smo{\R}{c}{c'} \geq 0$.

Additionally, we introduce the total score margin of $c$ as the sum of the score margin of $c$ w.r.t. all other candidates, $\movo{\R_i}{c} = \sum_{c' \neq c} \smo{\R_i}{c}{c'}$. Similarly, we have the total margin of victory of $c$, $\movo{\R}{c} = \sum_{c' \neq c} \smo{\R}{c}{c'}$.

Note that if $w \in \candidates$ is a Borda winner of $\R$ then $\movo{\R}{w} \geq 0$ since for all other candidates, $c$, $\smo{\R}{w}{c} \geq 0$.

Throughout the rest of this section, we will consider the margins of the winning candidate $w$. Hence, the dependency on $w$ of $\delta^w$ and $\Delta^w$ will be dropped.

We first introduce two useful lemmas.

\begin{lemma}\label{lem:wintop}
    Let $\R_i$ be a ballot of a rank matrix $\R = (\R_1,\dots,\R_\n)$ with $\n$ voters and $\m$ candidates s.t. $w \in \candidates$ is a Borda winner of $\R$.
    \begin{itemize}
        \item If $w \in \R_i$ and $\R_{i,1} = null$ then $$
        \R'_{i,k} =
        \left\{\begin{array}{cl}
        w & \text{if } k = 1 \\
        \R_{i,k} & \text{if } k \neq k_w \text{ with } k_w \text{ s.t. } \R_{i,k_w}=w
        \end{array}\right.
        $$ satisfies $\mov{\R_i'} > \mov{\R_i}$ and $|\R_i| = |\R'_i|$. 
        \item If $w \in \R_i$ and $\R_{i,1} \neq null$ then $$
        \R'_{i,k} =
        \left\{\begin{array}{cl}
        w & \text{if } k = 1 \\
        \R_{i,1} & \text{if }  k = k_w \text{ with } k_w \text{ s.t. } \R_{i,k_w}=w\\
        \R_{i,k} & \text{if } k \neq 1 \text{ and } k \neq k_w
        \end{array}\right.
        $$ satisfies $\mov{\R_i'} \geq \mov{\R_i}$ and $|\R_i| = |\R'_i|$. 
        \item If $w \not \in \R_i$ and $\R_{i,1} \neq null$ then $$
        \R'_{i,k} =
        \left\{\begin{array}{cl}
        w & \text{if } k = 1 \\
        \R_{i,k} & \text{if } k \neq 1
        \end{array}\right.
        $$ satisfies $\mov{\R_i'} > \mov{\R_i}$ and $|\R_i| = |\R'_i|$. 
    \end{itemize}
\end{lemma}

\begin{proof}
    If for all $c \in \candidates\setminus{\{w\}}$, $\sm{\R_i'}{c} \geq \sm{\R_i}{c}$ and there exists $c_0 \in \candidates\setminus{\{w\}}$, $\sm{\R_i'}{c_0} > \sm{\R_i}{c_0}$ then $\mov{\R_i'} > \mov{\R_i}$. 

    In the first case, for all $c \in dom(\R_i)\setminus{\{w\}}$, $\sigma^{max}_{\R_i'}(c) = \sigma^{max}_{\R_i}(c)$ and for all $c \not\in dom(\R_i)\setminus{\{w\}}$, $\sigma^{max}_{\R_i'}(c) < \sigma^{max}_{\R_i}(c)$ (the first entry, once free in $\R_i$ is not anymore in $\R_i'$). Additionally, $\sigma^{min}_{\R_i'}(w) > \sigma^{min}_{\R_i}(w)$. Hence, for all $c \in \candidates\setminus{\{w\}}$, $\sm{\R_i'}{c} > \sm{\R_i}{c}$ and $\mov{\R_i'} > \mov{\R_i}$.
    The second and third cases are analogous.
\end{proof}

We proved that in the optimal case for total margin of victory, if the first entry is locked it will only be with the winner and if the winner is locked it will only be in first entry. Now, let us turn our attention to the rest of the ranking.

\begin{lemma}\label{lem:loosebot}
    Let $\R_i$ be a ballot of a rank matrix $\R = (\R_1,\dots,\R_\n)$ with $\n$ voters and $\m$ candidates s.t. $w \in \candidates$ is a Borda winner of $\R$. If there exists $c \in \R_i\setminus{\{w\}}$ and $k_0,k_1 \in [|1,m|]$ s.t. $\R_{i,k_0} = null$, $\R_{i,k_1} = null$ and $k_0<k_c<k_1$ with $k_c$ s.t. $\R_{i,k_c}=c$ then $$
        \R'_{i,k} =
        \left\{\begin{array}{cl}
        c & \text{if } k = k_1\\
        \R_{i,k} & \text{if } k \neq k_c
        \end{array}\right.
        $$ satisfies $\mov{\R_i'} > \mov{\R_i}$ and $|\R_i| = |\R'_i|$. 
\end{lemma}

\begin{proof}
    Similarly to the previous proof, we compare for all $c' \in \candidates\setminus{\{w\}}$, $\sm{\R_i'}{c'}$ and $\sm{\R_i}{c'}$. If $w \in ran(\R_i)$, for all $c' \in \candidates\setminus{\{w,c\}}$, $\sm{\R_i'}{c'} = \sm{\R_i}{c'}$ and $\sm{\R_i'}{c} > \sm{\R_i}{c}$. Otherwise, for all $c' \in \candidates\setminus{\{w,c\}}$, $\sm{\R_i'}{c'} > \sm{\R_i}{c'}$ and $\sm{\R_i'}{c} > \sm{\R_i}{c}$ ($\sigma^{min}_{\R_i'}(w) > \sigma^{min}_{\R_i}(w)$). Finally, for all $c \in \candidates\setminus{\{w\}}$, $\sm{\R_i'}{c} > \sm{\R_i}{c}$. Hence, $\mov{\R_i'} > \mov{\R_i}$.
\end{proof}

Note that the previous lemma does not hold if there is no free entry before the locked entry . Indeed, otherwise, lowering the candidate entry will improve the first free entry and free candidates could end up with a better score margin. Hence, the presence of $k_0$.

We are ready to show that a specific pattern of locked entries dominate any other pattern of the same size.

\begin{definition}
    Let $\R_i$ be a ballot of a rank matrix $\R = (\R_1,\dots,\R_\n)$ with $\n$ voters and $\m$ candidates s.t. $w \in \candidates$ is a Borda winner of $\R$. If there exists $k_1,k_2\in [|1,\m|]$ such that $\R_{i,k} = null$ if and only if $k_1 < k < m+1-k_2$, we say that $\R_i$ is in \textbf{normal form} of parameter $k_1$ and $k_2$.
    When $k_1 + k_2 = \m$, we take $k_1 = \m$ and $k_2 = 0$.
\end{definition}

\begin{lemma}\label{th:normalform}
    Let $\R_i$ be a ballot of a rank matrix $\R = (\R_1,\dots,\R_\n)$ with $\n$ voters and $\m$ candidates s.t. $w \in \candidates$ is a Borda winner of $\R$. There exists $\R'_i$ in normal form of parameter $k_1$ and $k_2$ with $k_1+k_2=|\R_i|$ s.t. $\mov{\R_i'} \geq \mov{\R_i}$.
\end{lemma}

\begin{proof}
    For every ballot $ \R_i $, repeated applications of the transformations given in lemmas~\ref{lem:wintop} and~\ref{lem:loosebot} lead to the profile $ \R'_i $ with the given normal form.
\end{proof}

Now that we have introduced the normal form, it is possible to evaluate its total margin of victory.

\begin{lemma}\label{lem:normalformdelta}
    Let $\R_i$ be a ballot of a rank matrix $\R = (\R_1,\dots,\R_\n)$ with $\n$ voters and $\m$ candidates s.t. $w \in \candidates$ is a Borda winner of $\R$. Assuming there exists $k_1,k_2\in \mathbb{N}$ s.t. $\R_i$ is in normal form of parameter $k_1$ and $k_2$.
    \begin{itemize}
        \item If $k_1>0$, $\mov{\R_i} = (k_1 + k_2)\left(\m - \frac{(k_1 + k_2 + 1)}{2}\right)$.
        \item If $k_1=0$, $\mov{\R_i} = -(\m-1-k_2)^2 + \frac{k_2(k_2 + 1)}{2}$.
    \end{itemize}
\end{lemma}

\begin{proof}
    In the first case, $\mov{\R_i} = \sum_{c \neq w} \sm{\R_i}{c} = \sum_{i=1}^{k_1-1}i + (\m-k_1-k_2)k_1 + \sum_{i=\m-k_2}^{\m-1}i$ and in the second, $\mov{\R_i} = 0 - (\m-1-k_2)(\m-1-k_2) + \sum_{i=1}^{k_2}i$ where the first term is accounting for the candidates locked at the top of the ranking, the second, the free candidates and the third, the candidates locked at the bottom of the ranking.
\end{proof}

Finally, we can give an upper bound on the total margin of victory of any ballot.

\begin{lemma}\label{lem:maxcontrib}
    Let $\R_i$ be a ballot of a rank matrix $\R = (\R_1,\dots,\R_\n)$ with $\n$ voters and $\m$ candidates s.t. $w \in \candidates$ is a Borda winner of $\R$ and $|\R_i|\geq 1$
    $$\mov{\R_i} \leq (\m-1)(\m|\R_i|- (\m-1))$$
\end{lemma}

\begin{proof}
    Thanks to Theorem~\ref{th:normalform}, we know that ballots in normal form dominate other patterns in term of total margin of victory for a given number of locked entries. Thus, we just have to show that normal form ballots verify the inequality. 
    
    We derive the result from an upper bound on the score contribution per locked entry, $\frac{\mov{\R_i} - \mov{\varnothing}}{|\R_i|} \leq m(m-1)$ with $\mov{\varnothing} = - (m-1)^2$ the total margin of victory of the empty partial ballot.
    
    Let $\R_i$ be a ballot in normal form of parameter $k_1$ and $k_2$ s.t. $k_1 + k_2 \geq 1$.
    \begin{itemize}
        \item If $k_1>0$, $\frac{\Delta_{\R_i} - \Delta_{\varnothing}}{|\R_i|} = \frac{(m-1)^2}{k_1 + k_2} + m - \frac{(k_1 + k_2 + 1)}{2}$.\\
        Clearly, this is less than $m(m-1)$ with equality for $k_1=1$ and $k_2=0$.
        \item If $k_1=0$, $\frac{\Delta_{\R_i} - \Delta_{\varnothing}}{|\R_i|} = 2m - \frac{k_2}{2} - \frac{3}{2}$.\\
        Since $k_2\geq 1$, we have the desired result.
    \end{itemize}
    \vspace{-6mm}
\end{proof}

\begin{proof}[Proof of theorem~\ref{thm:lowerbound}]
    Let $\R$ be a rank matrix with $\n$ voters and $\m$ candidates s.t. $w \in \candidates$ is a Borda winner of $\R$.

    Assuming there exists $\mathcal{X}$, an AXp of size at most $n - \left \lfloor \frac{n}{m} \right \rfloor - 1$. Since $w\in \NW{\text{Borda}}{\mathcal{X}}$, we have $\mov{\mathcal{X}} \geq 0$.

    However, we have:

    \begin{eqnarray*}
    \mov{\mathcal{X}} 
    & = & \sum_{\mathcal{X}_i \in \mathcal{X}} \mov{\mathcal{X}_i}\\
    & \leq & \sum_{\mathcal{X}_i \in \mathcal{X}} (\m-1)(\m|\R_i|- (\m-1))\\
    & = & (\m-1)\bigg(\m\sum_{\mathcal{X}_i \in \mathcal{X}}|\R_i| - \sum_{\mathcal{X}_i \in \mathcal{X}}(\m-1)\bigg)\\
    & \leq & (\m-1)\left(\m\left(\n - \left \lfloor \frac{\n}{\m} \right \rfloor - 1 \right) - \n(\m-1)\right)\\
    & = & (\m-1)\left(\n - \left(\left \lfloor \frac{\n}{\m} \right \rfloor + 1\right)\m\right)\\
    & < & 0
    \end{eqnarray*}
    Contradiction.
    Hence, no AXp of size at most $n - \left \lfloor \frac{n}{m} \right \rfloor - 1$ exists for any rank matrix.
\end{proof}
    
\end{toappendix}

We are then able to show that the lower bound of Theorem~\ref{thm:lowerbound} is tight: let $\mathcal{X} = (\mathcal{X}_1,\dots,\mathcal{X}_\n)$ be the partial rank matrix where every entry is $null$ except that $\mathcal{X}_{i,1} = w$ for $ 1 \leq i \leq n - \left\lfloor \frac{\n}{\m} \right \rfloor $. 
It is not difficult to check that the minimum possible score for $ w $ is always greater than, or equal to, the maximum possible score for all other candidates. By Theorem~\ref{thm:lowerbound}, $\X$ is minimal and therefore it is an AXp  for any complete profile $ \R $ that extends it. 

\begin{theoremrep}\label{thm:equality}
    Let $\R$ be a rank matrix with $\n$ voters and $\m$ candidates and $w \in \candidates$ a Borda winner of $\R$ s.t. $\mathcal{X} \subseteq \R$ where $\mathcal{X}$ is the partial rank matrix where $w$ is ranked first for $n - \left \lfloor \frac{n}{m} \right \rfloor$ ballots and every other entry is $null$.
    $\mathcal{X}$ is an AXp of $\R$ and $|\X| = n - \left \lfloor \frac{n}{m} \right \rfloor$.
\end{theoremrep}

What can be concluded from Theorems~\ref{thm:lowerbound} and \ref{thm:equality} is that even the smallest AXps of a preference profile are rather long to be considered compact explanations, at least for large voter populations---the bound is linear in the number of voters---opening an interesting challenge of identifying feature spaces that result in more compact explanations. In Section~\ref{sec:discussion} we discuss this point in more details.

\begin{appendixproof}
    Let $\mathcal{X} = (\mathcal{X}_1,\dots,\mathcal{X}_\n)$ be the partial rank matrix where every entry is $null$ except for $i\in [|1,\n - \left \lfloor \frac{\n}{\m} \right \rfloor|]$ where $\mathcal{X}_{i,1} = w$.
    For all $c\in \candidates\setminus{\{w\}}$, we have:
    \begin{eqnarray*}
    \sm{\mathcal{X}}{c} & = &
    \sum_{\mathcal{X}_i \in \mathcal{X}} \sm{\mathcal{X}_i}{c}\\
    & = & \sum_{i=1}^{n - \left \lfloor \frac{n}{m} \right \rfloor} \sm{\mathcal{X}_i}{c} + \sum_{i=n - \left \lfloor \frac{n}{m} \right \rfloor +1}^{n}\sm{\mathcal{X}_i'}{c} \\
    & = & \left(n - \left \lfloor \frac{n}{m} \right \rfloor\right)1 - \left \lfloor \frac{n}{m} \right \rfloor (m-1) \\
    & = & n - \left \lfloor \frac{n}{m} \right \rfloor m\\
    & \geq & 0
    \end{eqnarray*}
    Hence, $w\in \NW{\text{Borda}}{\mathcal{X}}$. Since there is no smaller AXp (Theroem~\ref{thm:lowerbound}), $\mathcal{X}$ is an AXp of $\R$ and $n - \left \lfloor \frac{n}{m} \right \rfloor$ locked entries suffice.
\end{appendixproof}


\subsection{A map of elections for \shortAXps ~for Borda}

To get a clearer picture of the sizes of \shortAXps ~of the Borda rule we set up a map of elections using the tools by~\citet{szufa2020drawing}. 
A map of elections starts from a set of profiles, generated using known preference distributions, and generates a 2D embedding of the space obtained by calculating the isomorphic swap-distance between the profiles. 
Following their approach, to improve the interpretability of the map we introduce specific extreme profiles to act as a compass~\citep{ijcai2021p9,ijcai2023p299}. 
Our data set is composed of 146 profiles with 4 candidates and 12 voters generated from 17 different cultures. The details of our dataset can be found in the extended version of this paper~\citep{extended}. Preliminary experiments with higher number of voters and candidates led to excessive execution time for Algorithm~\ref{alg:findSiAXp}.

\begin{toappendix}
\section{Data set composition of experiment on \shortAXps}
\begin{tabular}{r l c}
    model & parameter & \#profiles \\ \hline
    Impartial Culture & & 10\\
    Impartial Anonymous Culture & & 10\\
    normalized Mallows & $\phi =0.5$ & 10 \\
    Urn model & $\alpha = 0.1$ & 10 \\ \hline
    Single-Peaked (Conitzer) & & 10 \\
    Single-Seaked (Walsh) & & 10\\
    Single-Peaked On a Circle & &10\\
    Single-Crossing & & 10\\
    Group-Separable & & 10\\ \hline
    1-Cube (Interval) & uniform interval & 10\\
    3-Cube (Cube) & uniform cube & 10\\
    Disc & disc in 2D & 10\\
    Circle & circle in 2D & 10\\
    Sphere & sphere in 3D & 10\\ \hline
    Uniformity (UN) & & 4\\
    Identity (ID) & & 1\\
    Antagonism (AN) & & 1\\ \hline
\end{tabular}
\end{toappendix}

\begin{figure*}[t]
    \centering
    \begin{minipage}{0.85\textwidth}
    \includegraphics[width=\linewidth]{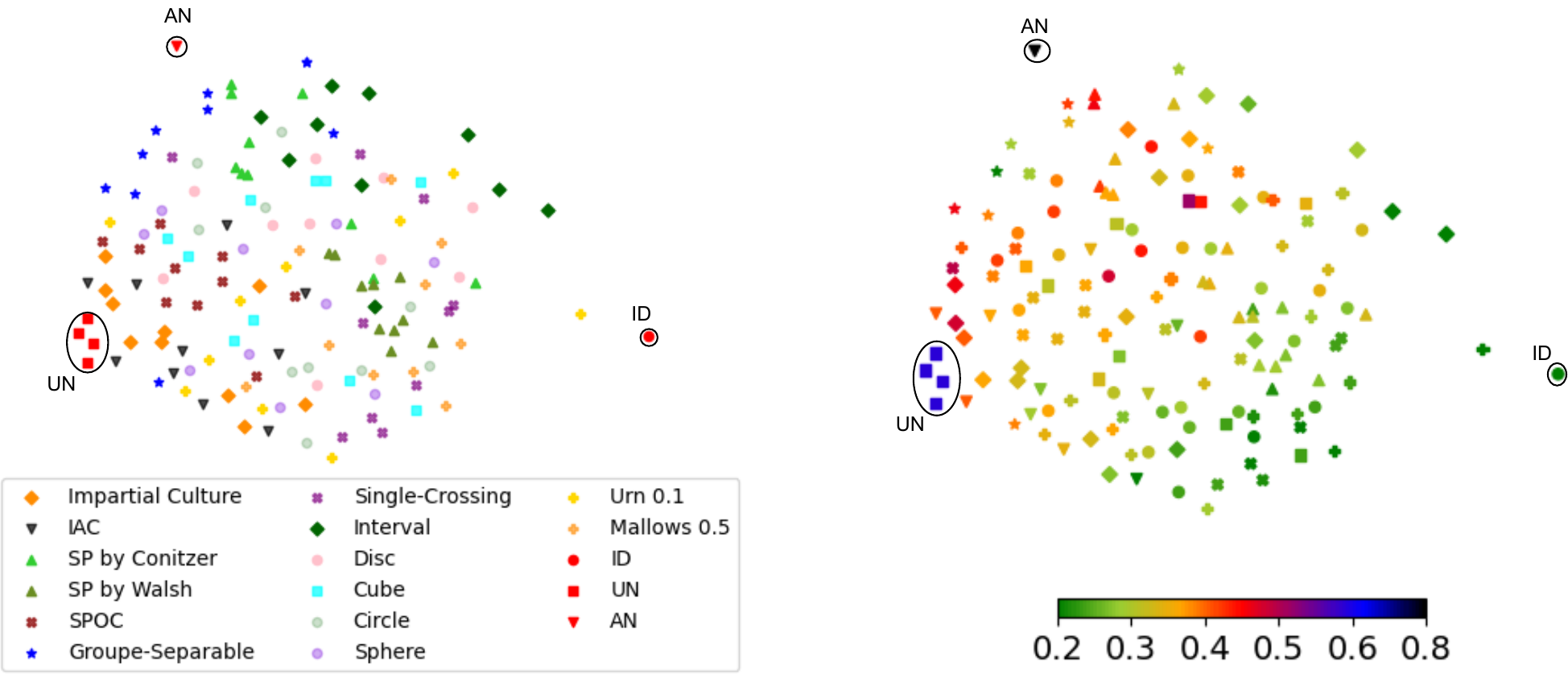}
    \caption{Map of elections for 146 preference profiles generated by fourteen different cultures plus six compass profiles (AN, ID, and four for UN). The map on the right shows for each preference profile the size of its \shortAXp, normalised as $\frac{|\shortAXp|}{\n\m}$.}
    \label{fig:2dembedding}
    \end{minipage}
\end{figure*}

\bigskip

The first map in Figure~\ref{fig:2dembedding} plots the map of elections for our dataset on the left, which is coherent with those obtained by \citet{szufa2020drawing} in past research.
The compass profiles in red are the identical culture (ID) where all ballots are identical, the antagonism culture (AN) where half of the ballots are identical and the other half is their opposite, and the uniform culture (UN) where ballots are drawn uniformally at random among all the possible ones (there are four of them to account for randomness). Ten instances were generated for each of the remaining fourteen cultures. For a detailed description of statistical cultures for preference profile generation used by maps of elections we refer to the presentation by \citet{szufa2020drawing}.

The second map in Figure~\ref{fig:2dembedding} shows the size of the \shortAXp ~for each of the profile in the dataset. 
We can observe that the antagonism and uniform profiles have the largest \shortAXp ~in our data set, and the identical profile has the smallest one (in coherence with our Theorem~\ref{thm:equality}).
When considering prefence distributions, we can observe that impartial culture (IC) generates profiles with long \shortAXps, given their closeness to uniform profiles. A similar observation can be drawn for the impartial anonymous culture (IAC) and single-peaked on a circle (SPOC).
A surprising observation is that the generation of single-peaked profiles have a strong impact on the size of \shortAXps, as can be seen by the Conitzer-SP profiles which have large \shortAXps ~while the Walsh-SP profiles have short \shortAXps. 
The well-known Urn and Mallow models (plotted here for one specific parameter only), have relatively small \shortAXps ~with some exceptions.



\begin{figure}[h]
    \centering
    \includegraphics[width=\linewidth]{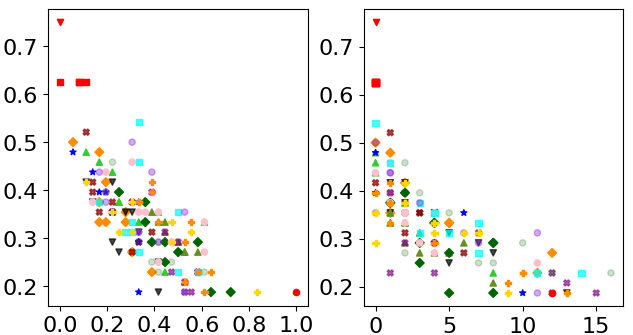}
    \caption{Normalised SiAXp size on the $x$-axis compared with agreement index (left) and margin of victory (right) 
    for the 146 profiles in our dataset.}
    \label{fig:SiAXpsie_corr}
\end{figure}

\bigskip

To investigate closely which properties of a preference profile are indicators of a large \shortAXp, Figure~\ref{fig:SiAXpsie_corr} compares the size of \shortAXps ~with two measures: the \emph{margin of victory}, which is the difference in Borda score between the winner and the second-best candidate in a profile, and the (normalized) \emph{agreement index}, formally defined as $\sum_{a,b \in \candidates\text{, }a\neq b} \left|N_{a>b} - N_{b>a}\right|/\big(n\binom{|\candidates|}{2}\big)$ where $N_{a>b}$ is the number of voters preferring candidate $a$ to candidate $b$. The agreement index was studied recently by \citet{ijcai2023p299}, based on previous work by \citet{HashemiE14} and \citet{can2015measuring}.

We observe that the size of the SiAXp is negatively correlated in our dataset with both the agreement index and the margin of victory. Spearman test gives respectively a correlation coefficient of $-0.761$ (p-value smaller than $0.001$)
and $-0.828$ (again, p $<0.001$).
%
Both results have an intuitive interpretation.
Elections with a small margin of victory require more information to identify which of two candidates is the winner. 
Similarly, the less the voters agree the more the candidates' scores are similar, and the more information is needed to determine the winner of the election.


\section{Conclusions and future work}\label{sec:discussion}

A primary direction for future research is to test formal explanations on voting rules defined on the majority graph, devising a feature encoding that result in compact formal explanations.
While the axiomatic calculus devised by \citet{NardiEtAlAAMAS2022} leads to human-readable explanations supporting the choice of the winner in an election, our proposal of adapting formal explanations for scoring rules does not seem fit for this application, due to the large number of possible explanations and the size of smallest explanations, which is high as soon as the preference profile shows some complexity.

A surprising connection can be shown between formal explanations and bribery in voting, with CXps identifying the positions in the rankings that allow for a minimal destructive bribery attack, and AXps defining optimal protection against such attacks \citep{lu2020computational}.
%
Formal explanations in voting can also be used to study optimal preference elicitation strategies to compute the winner of a given election. This problem has been well-studied in the literature as the communication complexity \citep{ConitzerS05} and the sample complexity \citep{dey2015sample} of voting rules, with the size of smallest abductive explanations having a natural correspondence with the former concept. 

%





\newpage

\bibliography{mybibfile}

\cleardoublepage

\section*{Appendix}\label{section:appendix}

\appendix

\end{document}